\documentclass{article}
\usepackage{arxiv}
\usepackage[utf8]{inputenc}
\usepackage[T1]{fontenc}
\usepackage{hyperref} 
\usepackage{url}
\usepackage{booktabs}
\usepackage{amsfonts}
\usepackage{nicefrac}
\usepackage{microtype}

\usepackage{booktabs}
\usepackage{xcolor}
\usepackage{amsmath, amssymb}
\usepackage{graphicx}
\usepackage{subcaption}
\usepackage{adjustbox}
\usepackage{algorithmic}
\usepackage{algorithm}

\DeclareMathOperator{\E}{\mathbb{E}}

\begin{document}

\title{Jointly Pre-training with Supervised, Autoencoder, and Value Losses for Deep Reinforcement Learning}

\author{
  Gabriel V.~de la Cruz Jr., Yunshu Du and Matthew E.~Taylor \\
  School of Electrical Engineering and Computer Science\\
  Washington State University\\
  Pullman, WA 99164-2752 \\
  \texttt{\{gabriel.delacruz,yunshu.du,matthew.e.taylor\}@wsu.edu} \\
}
\maketitle

\begin{abstract}  
 Deep Reinforcement Learning (DRL) algorithms are known to be data inefficient. One reason is that a DRL agent learns both the feature and the policy \emph{tabula rasa}. Integrating prior knowledge into DRL algorithms is one way to improve learning efficiency since it helps to build helpful representations. In this work, we consider incorporating human knowledge to accelerate the asynchronous advantage actor-critic (A3C) algorithm by pre-training a small amount of non-expert human demonstrations. We leverage the supervised autoencoder framework and propose a novel pre-training strategy that jointly trains a weighted supervised classification loss, an unsupervised reconstruction loss, and an expected return loss. The resulting pre-trained model learns more useful features compared to independently training in supervised or unsupervised fashion. Our pre-training method drastically improved the learning performance of the A3C agent in Atari games of Pong and MsPacman, exceeding the performance of the state-of-the-art algorithms at a much smaller number of game interactions. 
 Our method is light-weight and easy to implement in a single machine. For reproducibility, our code is available at \url{github.com/gabrieledcjr/DeepRL/tree/A3C-ALA2019} 
\end{abstract}

%


\maketitle


\section{Introduction}
\emph{Deep Reinforcement Learning} (DRL) has been an increasingly popular general machine learning technique, significantly contributing to the resurgence in neural networks research. 
Not only can DRL allow machine learning algorithms to learn appropriate representations without extensive hand-crafting of input, it can also achieve record-setting performance across multiple types of problems \cite{dqn,a3c,silver2016mastering,silver2018general}. However, one of the main drawbacks of DRL is its data complexity. Similar to classic RL algorithms, DRL suffers from slow initial learning as it learns \emph{tabular rasa}. While acceptable in simulated environments, the long learning time of DRL has made it impractical for real-world problems where bad initial performance is unaffordable, such as in robotics, self-driving cars, and health care applications \cite{bojarski2017explaining, li2017deep, miotto2017deep}.

There are two components of learning in DRL: feature learning and policy learning. While DRL is able to directly extract features using a deep neural network as its nonlinear function approximator, this process adds additional training time on top of policy learning and consequently slows down DRL algorithms. In this work, we propose several \emph{pre-training} techniques to tackle the feature learning problem in DRL. We believe that by aiding one of the learning components, a DRL agent will be able to focus more on the policy learning thus improve the overall learning speed. 

Many techniques have been proposed to address the data inefficiency of DRL. Transfer learning has been shown to work well for RL problems \cite{taylor2009transfer}. The intuition is that knowledge acquired from previously learned \emph{source tasks} can be transferred to related \emph{target tasks} such that the target tasks learn faster since they are not learning from scratch. Learning from demonstrations (LfD) \cite{argall2009survey, dqfd, apexdqn, gao2018reinforcement} is also an effective way to accelerate learning. In particular, demonstration data of a task can be collected from either a human demonstrator or a pre-trained agent; a new agent can start with mimicking the demonstrator's behavior to obtain a reasonable initial policy quickly, and later on move away from the demonstrator and learns on its own. 
One can also leverage additional auxiliary losses to gather extra information about a task \cite{unreal, mirowski2016learning, schmitt2018kickstarting, du2018adapting}. For example, an agent can jointly optimize the policy loss and an unsupervised reconstruction loss; doing so explicitly encourages learning the features. In this work, we combine the flavor of the methods above and propose a pre-training strategy to speed up learning. Our method jointly pre-trains a supervised classification loss, an unsupervised reconstruction loss, and a value function loss. 


\section{Preliminaries}

\subsection{Deep Reinforcement Learning}
We consider a reinforcement learning (RL) problem that is modeled using a Markov Decision Process (MDP), represented by a 5-tuple $\langle S, A, P, R, \gamma \rangle$. A \emph{state} $S_{t}$ represents the environment at time $t$. An agent learns what \emph{action} $A_t \in \mathcal{A(}s\mathcal{)}$ to take in $S_{t}$ by interacting with the environment. A \emph{reward} $R_{t+1} \in \mathcal{R} \subset \mathbb{R}$ is given based on the action executed and the next state reached, $S_{t+1}$. The goal is to maximize the expected cumulative return $G_t = \sum_{k=0}^{\infty} \gamma^k R_{t+k+1}$, where $\gamma \in [0,1]$ is a discount factor that determines the relative importance of future and immediate rewards \cite{sutton2018}. 

In value-based RL algorithms, an agent learns the state-action value function $Q^{\pi}(s, a) = \E_{s'}[r+\gamma \max_{a'}Q^{\pi} (s', a')|s, a]$, and the optimal value function $Q^*(s,a) = max_{\pi}Q^{\pi}(s,a)$ gives the expected return for taking an action $a$ at state $s$ and thereafter following an optimal policy. However, directly computing Q values is not feasible when the state space is large. The deep Q-network (DQN) algorithm \cite{dqn} uses a deep neural network (parameterized as $\theta$) to approximate the Q function as $Q(s, a; \theta) \approx Q^*(s, a)$. At each iteration \emph{i}, DQN minimizes the loss 
\[
L_{i}(\theta_i) = \E_{s, a, r, s'} \Big[(y - Q(s, a; \theta_i))^2 \Big]
\]
where $y = r + \gamma max_{a'}Q(s', a';\theta_{i}^-)$ is the \emph{target network} (parameterized as $\theta_{i}^-$) that was generated from previous iterations. 
The key component that helps to stabilize learning is the \emph{experience replay memory} \cite{lin1993reinforcement} which stores past experiences. An update is performed by drawing a batch of 32 experiences (minibatch) uniformly random from the replay memory---doing so ensures the  $i.i.d.$ property of the sampled data thus stabilizes the learning. \emph{Reward clipping} also helps to make DQN work. All rewards are clipped to $[-1,1]$ thus avoids the potential instability brought by various reward scales in different environments. 

\subsection{Asynchronous Advantage Actor-Critic}
The asynchronous advantage actor-critic (A3C) algorithm \cite{a3c} is a policy-based method that combines the actor-critic framework with a deep neural network. A3C learns both a \emph{policy function} $\pi(a_t|s_t;\theta)$ (parameterized as $\theta$) and a \emph{value function} $V(s_t;\theta_v)$ (parameterized as $\theta_v$). The policy function is the \emph{actor} that decides which action to take while the value function is the \emph{critic} that evaluates the quality of the action and also bootstraps learning. The policy loss given by \citet{a3c} is
\begin{equation*}
\begin{split}
L^{a3c}_{policy} = & \nabla_\theta log(\pi(a_t|s_t;\theta))\bigr(Q^{(n)}(s_t, a_t;\theta,\theta_v)-V(s_t;\theta_v)\bigl) \\
& -\beta^{a3c}\mathcal{H}\nabla_\theta\bigl(\pi(s_t;\theta)\bigr)
\end{split}
\end{equation*}
where $Q^{(n)}(s_t, a_t;\theta,\theta_v)=\sum_{k=0}^{n-1} \gamma^{k}r_{t+k} + \gamma^{n}V(s_{t+n};\theta_v)$ is the $n$-step bootstrapped value that is bounded by a hyperparameter $t_{max}$ ($n\leq t_{max}$). $\mathcal{H}$ is an entropy regularizer for policy $\pi$ (weighted by $\beta^{a3c}$) which helps to prevent premature convergence to sub-optimal policies. The value loss is
\begin{equation*}
L^{a3c}_{value} = \nabla_{\theta_v}\Bigl(\bigl(Q^{(n)}(s_t, a_t;\theta,\theta_v)-V(s_t;\theta_v)\bigr)^2\Bigr)
\end{equation*}
The A3C loss is then 
\begin{equation} \label{eq:a3c_update}
L^{a3c}=L^{a3c}_{policy}+\alpha L^{a3c}_{value}
\end{equation}
where $\alpha$ is a weight for the value loss. A3C runs $k$ actor-learners in parallel and each with their own copies of the environment and parameters. An update is performed using data collected from all actors. In this work, we use the feed-forward version of A3C \cite{a3c} for all experiments. The architecture consists of three convolutional layers, one fully connected layer (\emph{fc1}), followed by two branches of a fully connected layer: a policy function output layer (\emph{fc2}) and a value function output layer (\emph{fc3}). 

\subsection{Transformed Bellman Operator for A3C}
While the reward clipping technique helped to reduce the variance and stabilize learning in DQN, \citet{dqfd} found that clipping all rewards to $[1,-1]$ hurts the performance in games where the reward has various scales. For example, in the game of MsPacman, a single dot is worth $10$ points, while a cherry bonus is worth $100$ points; when both are clipped to $1$, the agent becomes incapable of distinguishing between small and large rewards, resulting in reduced performance. \citet{apexdqn} proposed the \emph{transformed Bellman operator} to overcome this problem in DQN. Instead of changing the magnitude of rewards, \citet{apexdqn} considers reducing the scale of the action-value function, which enables DQN to use raw rewards instead of clipped ones. In particular, a transform function 
\begin{equation} \label{eq:h_function}
    h:z \mapsto sign(z)\left(\sqrt{|z| + 1} - 1\right) + \varepsilon z
\end{equation}
\noindent is applied to reduce the scale of $Q^{(n)}(s_t, a_t;\theta,\theta_v)$ and $Q$ is transformed as
\begin{equation} \label{eq:tb_q_function}
    Q^{(n)}_{TB}(s_t, a_t;\theta,\theta_v) = \sum_{k=0}^{n-1} h\left( \gamma^{k}r_{t+k} + \gamma^{n}h^{-1}\left( V\left(s_{t+n};\theta_v\right) \right) \right)
\end{equation}
In this work, we apply the transformed Bellman operator in the A3C algorithm and use the raw reward value (instead of clipped) to perform updates. We denote this method as \emph{A3CTB}.


\subsection{Self-Imitation Learning}\label{sec:sil}
The \emph{self-imitation learning} (SIL) algorithm aims to encourage the agent to learn from its own past good experiences \cite{sil}. Built on the actor-critic framework \cite{a3c}, SIL adds a replay buffer $\mathcal{D}={(s_t, a_t, G_t)}$ to store the agent's past experiences. The authors propose the following off-policy actor-critic loss
\begin{align}
    L^{sil}_{policy} &= - log(\pi(a_t|s_t;\theta)) \bigr(G_t-V(s_t;\theta_v)\bigl)_{+} \nonumber \\
    L^{sil}_{value} &= \frac{1}{2} ||\bigl(G_t-V(s_t;\theta_v)\bigr)_{+}||^2 \nonumber
\end{align}
where $(\cdot)_{+}=max(\cdot, 0)$ and $G_t=\sum_{k=0}^{\infty}\gamma^{k}R_{t+k+1}$ is the discounted sum of rewards. The SIL loss is then
\begin{equation}\label{eq:sil_update}
    L^{sil} = L^{sil}_{policy}+\beta^{sil}L^{sil}_{value}
\end{equation}
In this work, we leverage this framework and incorporate SIL in \emph{A3CTB} (see Section \ref{sec:a3csil}).


\subsection{Supervised Pre-training}\label{sec:slpretrain}
In our previous work, supervised pre-training consists of a two-stage learning hierarchy~\cite{de2018pre}: 1) pre-training on human demonstration data, and 2) initializing a DRL agent's network with the pre-trained network $\theta \gets \theta_{s}$. It uses non-expert human demonstrations as its training data where the game states are the neural network's inputs and assume the non-optimal human actions as the true labels for each game state. The network is pre-trained with the cross-entropy loss
\begin{equation*}
    L_s = - \sum_{\forall x} p(x) log\bigl(q(x; \theta_{s})\bigr)
\end{equation*}
where $x$ is the image game state $s$ and $p(x)$ is the distribution over discrete variable $x$ represented here as a one-hot vector of the human action $a$; while $q(x; \theta_{s})$ is the output distribution of the supervised learning neural network with the weights $\theta_s$.

\subsection{Supervised Autoencoder}
An autoencoder learns to reconstruct its inputs and has been used for unsupervised learning of features. A supervised autoencoder (SAE) is an autoencoder with an additional supervised loss that can better extract representations that are tailored to the class labels. 
For example in \citet{sae}, the authors consider a supervised learning setting where the goal is to learn a mapping between some inputs $X \in \mathbb{R}$ and some targets $Y \in \mathbb{R}$. Instead of learning with only a supervised loss, an auxiliary reconstruction loss is integrated 
and the following SAE objective is proposed: 
\begin{equation}
    L^{sae}=L^{sae}_{s}(W_{s}Fx_i,y_i) + L^{sae}_{ae}(W_{ae}Fx_i, x_i)
\end{equation}
where $F$ is the weights of a neural network; $W_s$ and $W_{ae}$ are the weights for the supervised output layer and the autoencoder output layer respectively. Here, $L^{sae}_{s}$ and $L^{sae}_{ae}$ can be any loss functions (e.g., MSE). 

Existing work have considered training using an SAE loss from scratch. Our method is different in that i) we consider the SAE loss as a pre-training method instead of training from scratch and ii) we jointly pre-train a supervised loss and a reconstruction loss and then use the learned parameters as initialization (i.e., the two-stage hierarchy described in Section \ref{sec:slpretrain}). In this work, we explore if incorporating an unsupervised loss in supervised pre-training can further boost the learning of an agent. 


\section{Methodologies}\label{sec:method}
This section describes our proposed algorithms. First, we show that by incorporating the SIL framework (see Section \ref{sec:sil}), we can further improve the performance of the original A3C algorithm \cite{a3c} and the \emph{A3CTB} variant from our previous work \cite{de2018pre}; we term this new method as \emph{A3CTB+SIL}. Then, we introduce our proposed pre-training methods and show that, after pre-training, \emph{A3CTB+SIL} can achieve superior results; its performance on MsPacman exceeds or is comparable to some state-of-the-art algorithms that use human demonstrations (e.g., \citet{dqfd,sil}), and is also much lower on computational demands.   


\subsection{A3C with Self-Imitation Learning}\label{sec:a3csil}
We incorporate the self-imitation learning (SIL) framework (see Section \ref{sec:sil}) in \emph{A3CTB} with the following modifications. To enable using raw rewards (as was done in \emph{A3CTB}), 
we apply the transformation function $h$ (Equation~\eqref{eq:h_function})
to the returns as $G_t=h\bigr(r_{t+1} + \gamma h^{-1}(G_{t+1})\bigl)$. 
We also add a SIL-learner in parallel with the $k$ actor-learners in A3C (i.e., there are a total of $k+1$ parallel threads). 
The SIL-learner does not have its own copy of the environment; it learns by optimizing $L^{sil}$ using minibatch sampling from $\mathcal{D}$. The SIL-learner acts similarly to the other actor-learners as it updates the global network asynchronously. Each actor-learner contributes to $\mathcal{D}$ through a shared episode queue $\mathcal{Q}_{\mathcal{E}}$, where 
$\mathcal{E}$ is an episode buffer for each actor-learner that stores observation at time $t$ as $\{s_t,a_t,r_t\}$, until a terminal state is reached (i.e., the end of an episode). At a terminal state, the actor-learner computes the returns, $G_t$, with the transformation, $h$, for each step in the episode. Then $\mathcal{E}$ with the computed transformed returns are added to the shared episodes queue $\mathcal{Q}_{\mathcal{E}}$. The pseudocode for the SIL-learner is shown in Algorithm~\ref{alg:a3ctbsil_silworker}. We denote this method as \emph{A3CTB+SIL}.

\begin{algorithm}[ht!]
    \caption{SIL-learner in A3CTB+SIL.}
    \label{alg:a3ctbsil_silworker}
    \begin{algorithmic}[1]
        \raggedright
        \STATE \textit{// Assume global shared parameter vector $\theta$ and $\theta_v$ and global shared counter $T=0$} \\
        \STATE \textit{// Assume global shared episodes queue $\mathcal{Q}_{\mathcal{E}}$} \\
        \STATE \textit{// Assume thread-specific parameter vectors $\theta^{'}$ and $\theta_{v}^{'}$} \\
        \STATE Initialize replay memory $\mathcal{D} = \emptyset$ \\
        \REPEAT
            \STATE Synchronize parameters $\theta^{'} \gets \theta$ and $\theta_{v}^{'} \gets \theta_v$ \\
            \FOR{$m \gets 1$ \TO $M$}
                \STATE Sample a minibatch $\{s_j, a_j, G_j\}$ from $\mathcal{D}$\\
                \STATE Compute gradients w.r.t. $\theta^{'}: d\theta \gets \nabla_{\theta^{'}} \text{log}\pi(a_j|s_j;\theta^{'})(G_j-V(s_j;\theta_{v}^{'}))_{+}$ \\
                \STATE Compute gradients w.r.t. $\theta_{v}^{'}: d\theta_v \gets \partial((G_j-V(s_j;\theta_{v}^{'}))_{+})^2 / \partial\theta_{v}^{'}$ \\
                \STATE Perform asynchronous update of $\theta$ using $d\theta$ and $\theta_v$ using $d\theta_v$ \\
            \ENDFOR
            \WHILE{\texttt{len($\mathcal{Q}_{\mathcal{E}}$)} $> 0$}
                \STATE Dequeue first episode $\mathcal{E}$ from $\mathcal{Q}_{\mathcal{E}}$ \\
                \STATE $\mathcal{D} \gets \mathcal{D} \cup \{S_t, A_t, G_t\}$ for all $t$ in $\mathcal{E}$ \\
            \ENDWHILE
        \UNTIL{$T > T_{max}$}
    \end{algorithmic}
\end{algorithm}

\subsection{Pre-training Methods}

We now introduce our pre-training methods. The same set of non-expert human demonstration data collected from \citet{de2018pre} is used for all pre-training (see Table~\ref{table:human_demo}). This work deviates from our previous work in two aspects: 
1) we integrate multiple losses for pre-training while the previous work only considered supervised pre-training, and 2) we train using \emph{A3CTB+SIL} while the previous work train using \emph{A3CTB}; 
we shall see that integrating SIL is beneficial for our pre-training approach and will discuss the reasons later in this section. 

From our previous work, supervised pre-training on demonstrations can accelerate learning in A3C and \emph{A3CTB}~\cite{de2018pre}; for brevity, we denote this method as \emph{[SL]}\footnote{We denote all pre-training methods with their names in brackets.} pre-training.
However, there were two potential problems with \emph{[SL]} pre-training: 1) the value output layer (fc3) of A3C is not pre-trained, and 2) training only to minimize the loss between human and agent actions is sub-optimal since the action demonstrated by the human could be noisy.


To tackle the first problem, we pre-train an aggregated loss that consists of the supervised and the value return losses. For the supervised component, we use the SIL policy loss $L^{sil}_{policy}$ which can be interpreted as the cross-entropy loss $-log(\pi(a_t|s_t;\theta))$ weighted by $(G_t-V(s_t;\theta_v))_+$~\cite{sil}. The $(\cdot)_+$ operator encourages the agent to imitate past decisions only when the returns are larger than the value. The value return loss is nearly identical to the value loss $L^{sil}_{value}$ in SIL, but without the $(\cdot)_+$ operator. Note that this is also similar to A3C's value loss $L^{a3c}_{value}$, but instead of the $n$-step bootstrapped value $Q^{(n)}$, we use the discounted returns $G_t$, which can be easily computed from the human demonstration data since it contains the full trajectory of each episode. 
We denote this method as \emph{[SL+V]} pre-training.

\begin{figure}[t!]
    \centering
    \includegraphics[width=0.75\textwidth]{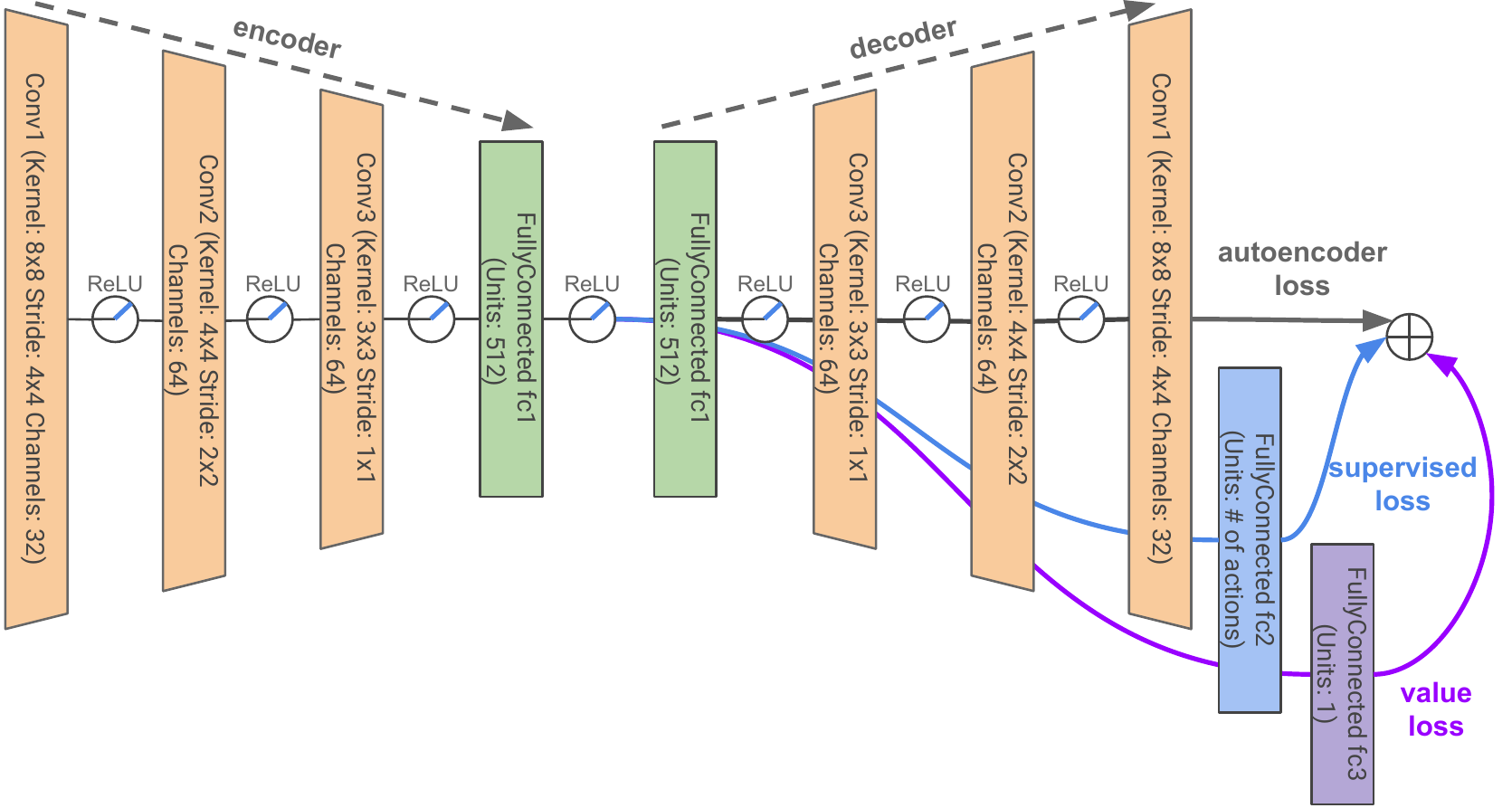}
    \caption{Network architecture on pre-training with a combined loss of supervised, value and autoencoder losses. 
    }
    \label{fig:sae_v_network}
\end{figure}

\citet{de2018pre} also revealed that supervised pre-training learns features 
that are geared more towards the supervised loss. For example in Pong, the area around the paddle is an important region since the paddle movements are associated with 
human actions. This implies the second problem mentioned above; if the features learned to focus on human actions only, they might not generalize well to new trajectories. To obtain extra information in addition to the supervised features, we take inspiration from the supervised autoencoder framework which jointly trains a classifier and an autoencoder~\cite{sae}; we believe this approach will retain the important features learned through supervised pre-training and at the same time, learns additional general features from the added autoencoder loss. 
Finally, we blend in the value loss $L^{saev}_{v}$ with the supervised and autoencoder losses as 
\begin{align}\label{eq:saev_loss}
    L^{saev}=L^{saev}_{s}(W_{s}Fx_i,y_i) &+ L^{saev}_{ae}(W_{ae}Fx_i, x_i) \nonumber \\
    &+ L^{saev}_{v}(W_{v}Fx_i, x_i).
\end{align}
\noindent We denote this method as \emph{[SL+V+AE]} pre-training. The network architecture of this pre-training method is shown in Figure~\ref{fig:sae_v_network}. In this network, Tensorflow's \emph{SAME} padding option is used to ensure that the input size is the same as the output size which is inherently necessary for reconstructing the input. 
This change results in a final output of $88x88$ of the neural network due to the existing network architecture filters used. Thus, instead of downsizing the output from $88x88$ to $84x84$ (which is the original input size of A3C), we changed the input size to $88x88$ in the spirit of the work of~\citet{kimura2018daqn} which uses autoencoder for pre-training.

There are two strong motivations to use self-imitation learning in A3C when using pre-training. First, human demonstration data can be loaded into SIL's memory to jumpstart the memory and continue learning with the data. Second, the motivation of jointly pre-train with multiple losses, especially with a value loss, is not only to learn better features but also to use the pre-trained policy and value layers into the A3C network. Adding the value loss allows pre-training the entire network. In turn, since SIL self-imitates its own experience, data generated during early stages are potentially 
closely related to the policy and value learned from pre-training; 
the learning speed could be increased more at the early stage of training. By using SIL, pre-training addresses feature learning while implicitly addressing policy learning.

\begin{table}[pt!]
    \caption{Human demonstration size and quality, collected from \citet{de2018pre}.}
    \label{table:human_demo}
    \centering
    \adjustbox{max width=0.5\textwidth}{
        \begin{tabular}{r|r|r|r|r}
        Game          & Worst score & Best score    & \# of states & \# of episodes \\ \hline \hline
        MsPacman      & 4020        & 18241         & 14504        &  8  \\ \hline
        Pong          & -13         & $\mathbf{5}$  & 21674        &  6  \\ \hline
        \end{tabular}
    }
\end{table}

\begin{table}[th]
    \caption{All games use the same set of hyperparameters except for Pong, where we found setting RMSProp epsilon to $1\times10^{-4}$ gives a much more stable learning.}
    \label{table:hyperparameters}
    \centering
    \adjustbox{max width=\textwidth}{
        \begin{tabular}{r|r}
        \textbf{Common Parameters}        & \textbf{Value} \\ \hline \hline
        Input size             & 88$\times$88$\times$4 \\ \hline
        Padding method         & SAME     \\ \hline \hline
        \multicolumn{2}{c}{\textbf{Parameters unique to pre-training}} \\ \hline \hline
        Adam learning rate &  $5\times 10^{-4}$ \\ \hline
        Adam epsilon &  $1\times 10^{-5}$ \\ \hline
        Adam $\beta_1$ &  $0.9$ \\ \hline
        Adam $\beta_2$ &  $0.999$ \\ \hline
        L2 regularization weight & $1\times 10^{-5}$     \\ \hline
        Number of minibatch updates & 50,000     \\ \hline
        Batch size             & 32     \\ \hline \hline
        \multicolumn{2}{c}{\textbf{Parameters unique to A3C}} \\ \hline \hline
        RMSProp learning rate  & $7 \times 10^{-4}$  \\ \hline
        RMSProp epsilon        & $1 \times 10^{-5}$  \\ \hline
        RMSProp decay          & 0.99      \\ \hline 
        RMSProp momentum       & 0      \\ \hline 
        Maximum gradient norm  & 0.5     \\ \hline
        $k$ parallel actors  & 16      \\ \hline
        $t_{max}$  & 20          \\ \hline
        transformed Bellman operator $\varepsilon$ & $10^{-2}$ \\ \hline 
        \multicolumn{2}{c}{\textbf{Parameters unique to SIL}} \\ \hline \hline
        M & 4 \\ \hline
        $\beta^{sil}$ & $0.5$ \\ \hline
        replay buffer $\mathcal{D}$ size & $10^6$\\ \hline \hline
        \end{tabular}
    }
\end{table}

\section{Experiments and Results}\label{sec:result}
We then present our experiments. First, we show that \emph{A3CTB+SIL} exceeds the performance of A3C and \emph{A3CTB}. 
As a comparison, we also evaluate \emph{A3C+SIL} since \citet{sil} implemented SIL in the \emph{synchronous} version of A3C (i.e., A2C) \cite{a3c}, which is different from our implementation that the SIL-learner is \emph{asynchronous}. 
Then, we present our experiments and results for the pre-training approaches (all trained in \emph{A3CTB+SIL} after pre-training) and show that they all outperform the baseline A3C algorithm. 
We use the same set of parameters across all experiments, shown in Table \ref{table:hyperparameters}.

\begin{figure*}[ht]
    \centering
    \begin{subfigure}[b]{0.49\textwidth}
        \includegraphics[width=\textwidth]{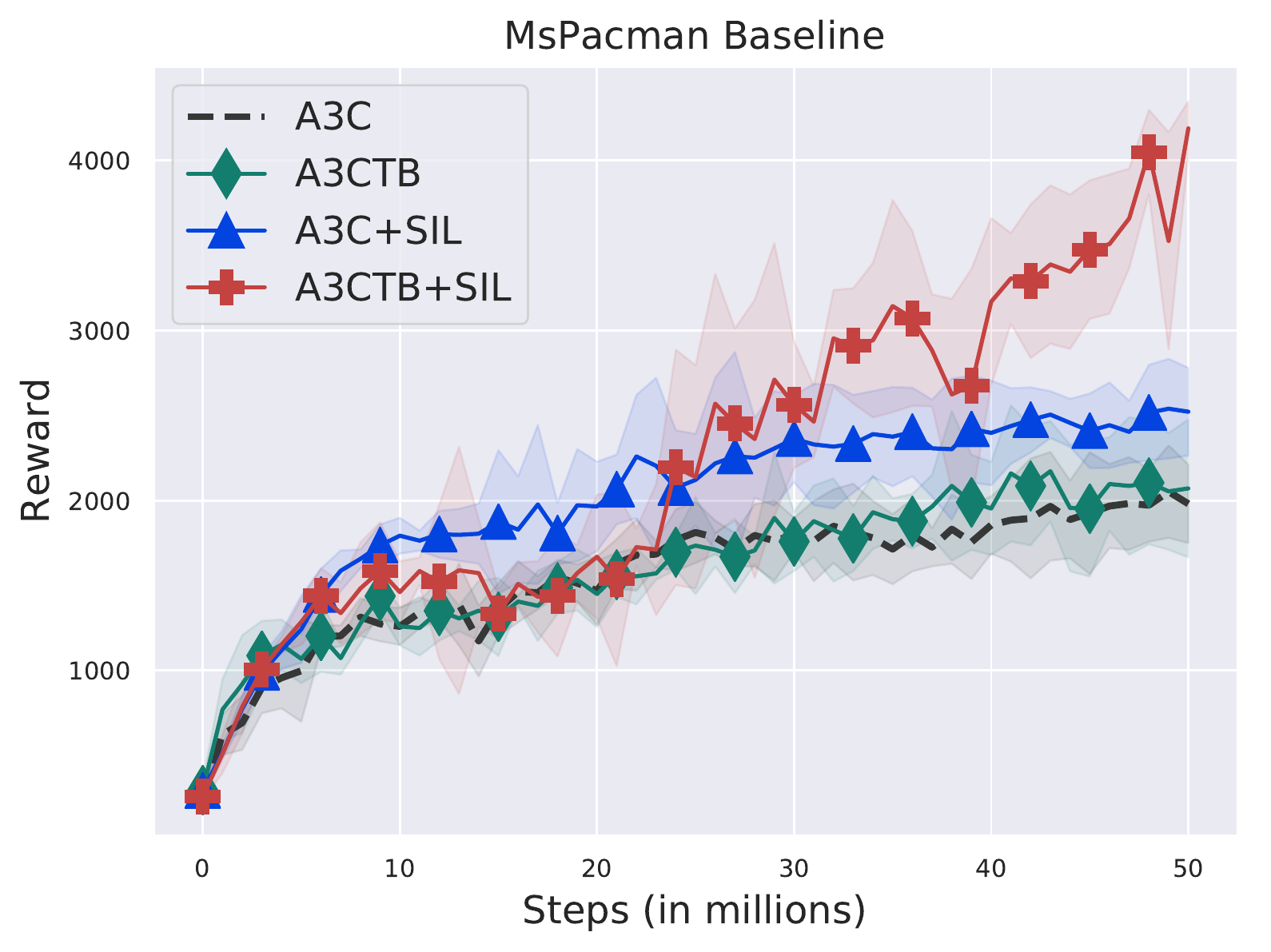}
        \label{fig:pacman-base}
    \end{subfigure}
    \begin{subfigure}[b]{0.49\textwidth}
        \includegraphics[width=\textwidth]{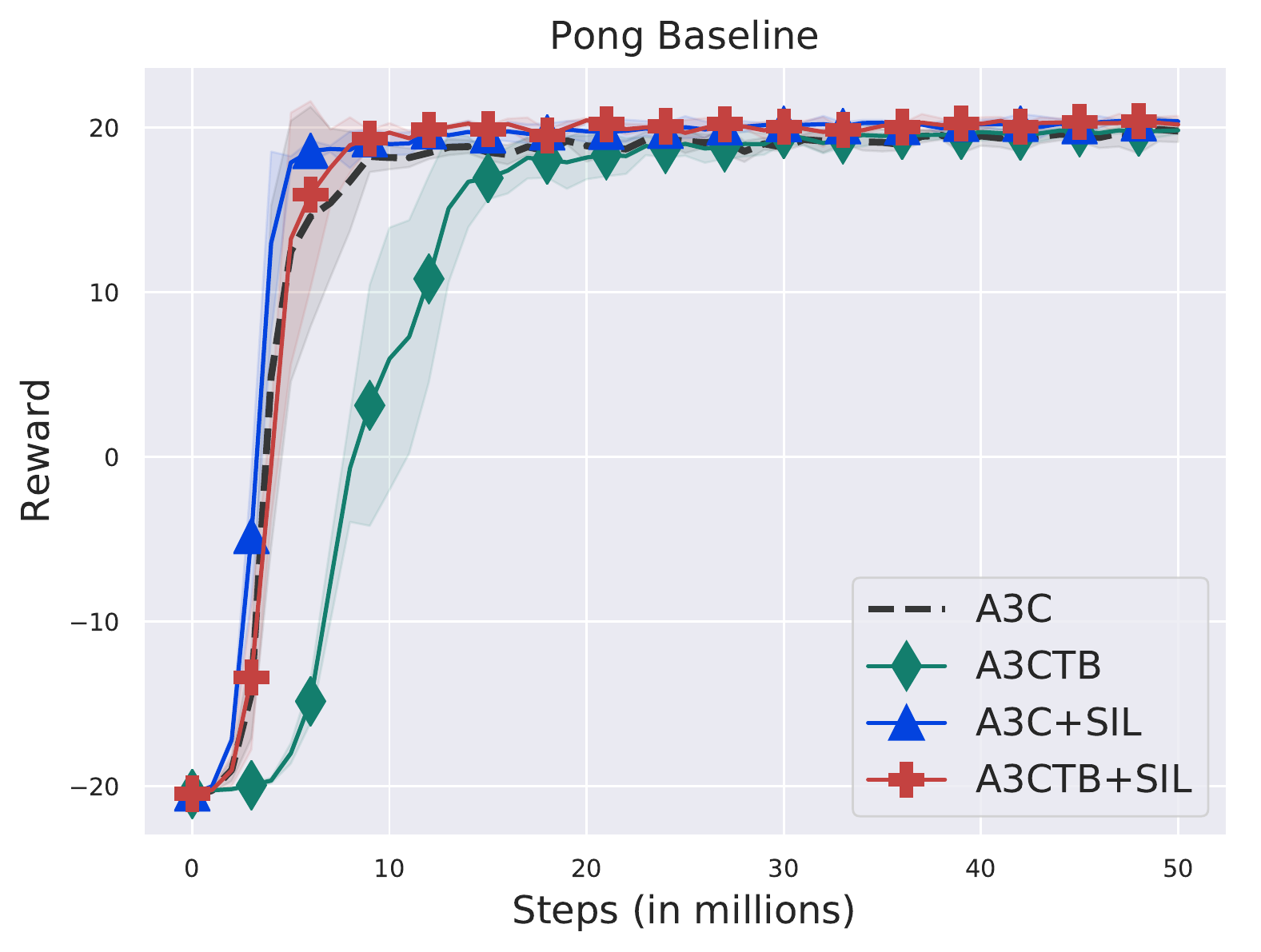}
        \label{fig:pong-base}
    \end{subfigure}
    \caption{Baseline performance of MsPacman and Pong without pre-training. The x-axis is the total number of training steps (16 actors for methods without SIL; 16 actors plus 1 SIL-learner for methods using SIL). Each step consists of four game frames (frame skip of four). The y-axis is the average testing score over three trials; shaded regions are the standard deviation.}
    \label{fig:base}
\end{figure*}

\begin{figure*}[ht]
    \centering
    \begin{subfigure}[b]{0.49\textwidth}
        \includegraphics[width=\textwidth]{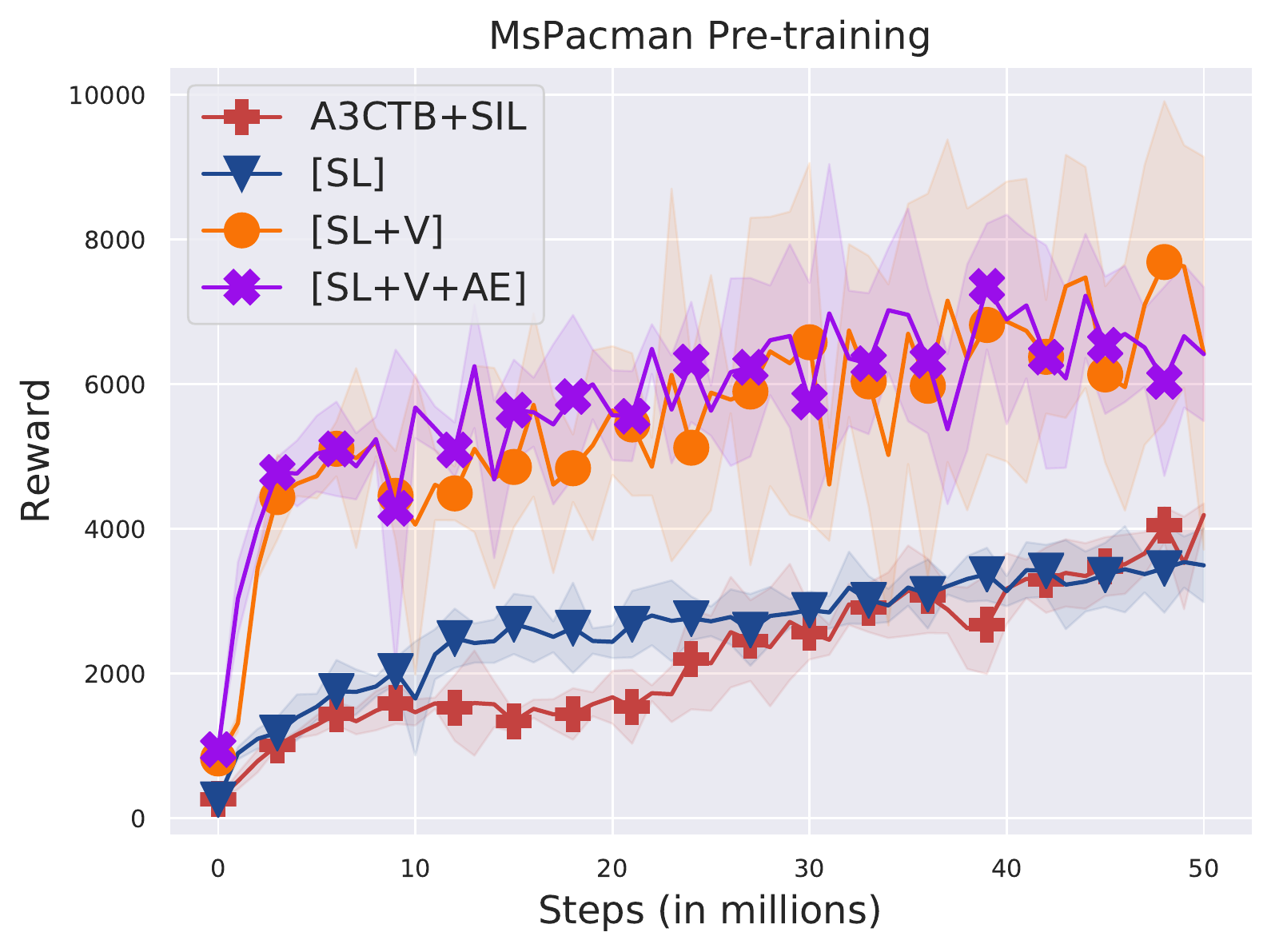}
    \end{subfigure}
    \begin{subfigure}[b]{0.49\textwidth}
        \includegraphics[width=\textwidth]{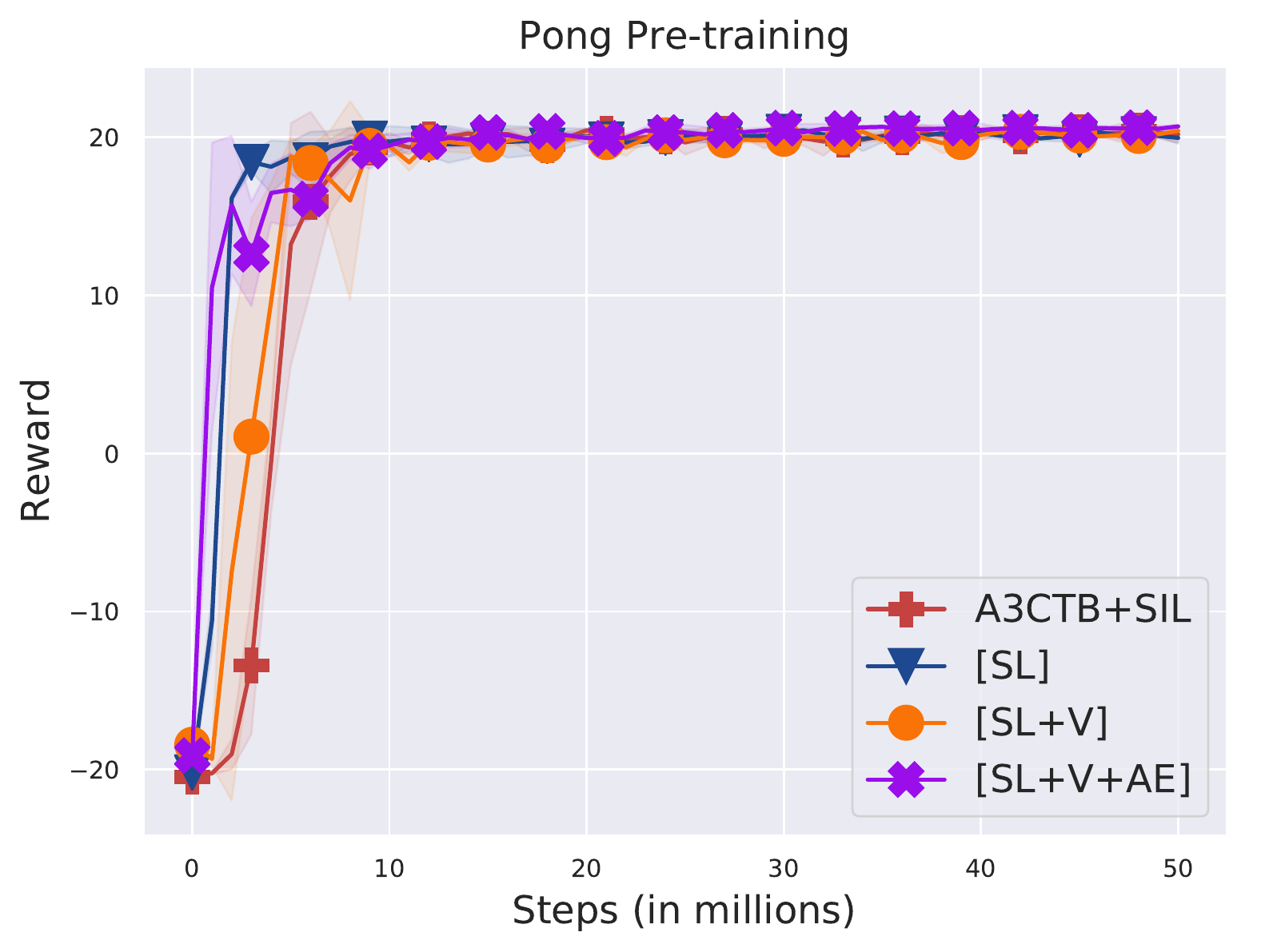}
    \end{subfigure}
    \caption{Pre-training performance of MsPacman and Pong. All layers are transferred. Pre-training methods are shown in brackets. After pre-training, all agents are trained in \emph{A3CTB+SIL}. The x-axis is the total number of training steps (16 actors for methods without SIL; 16 actors plus 1 SIL-learner for methods using SIL). Each step consists of four game frames (frame skip of four). The y-axis is the average testing score over three trials; shaded regions are the standard deviation.}
    \label{fig:transall}
\end{figure*}

\begin{figure*}[ht]
    \centering
    \begin{subfigure}[b]{0.49\textwidth}
        \includegraphics[width=\textwidth]{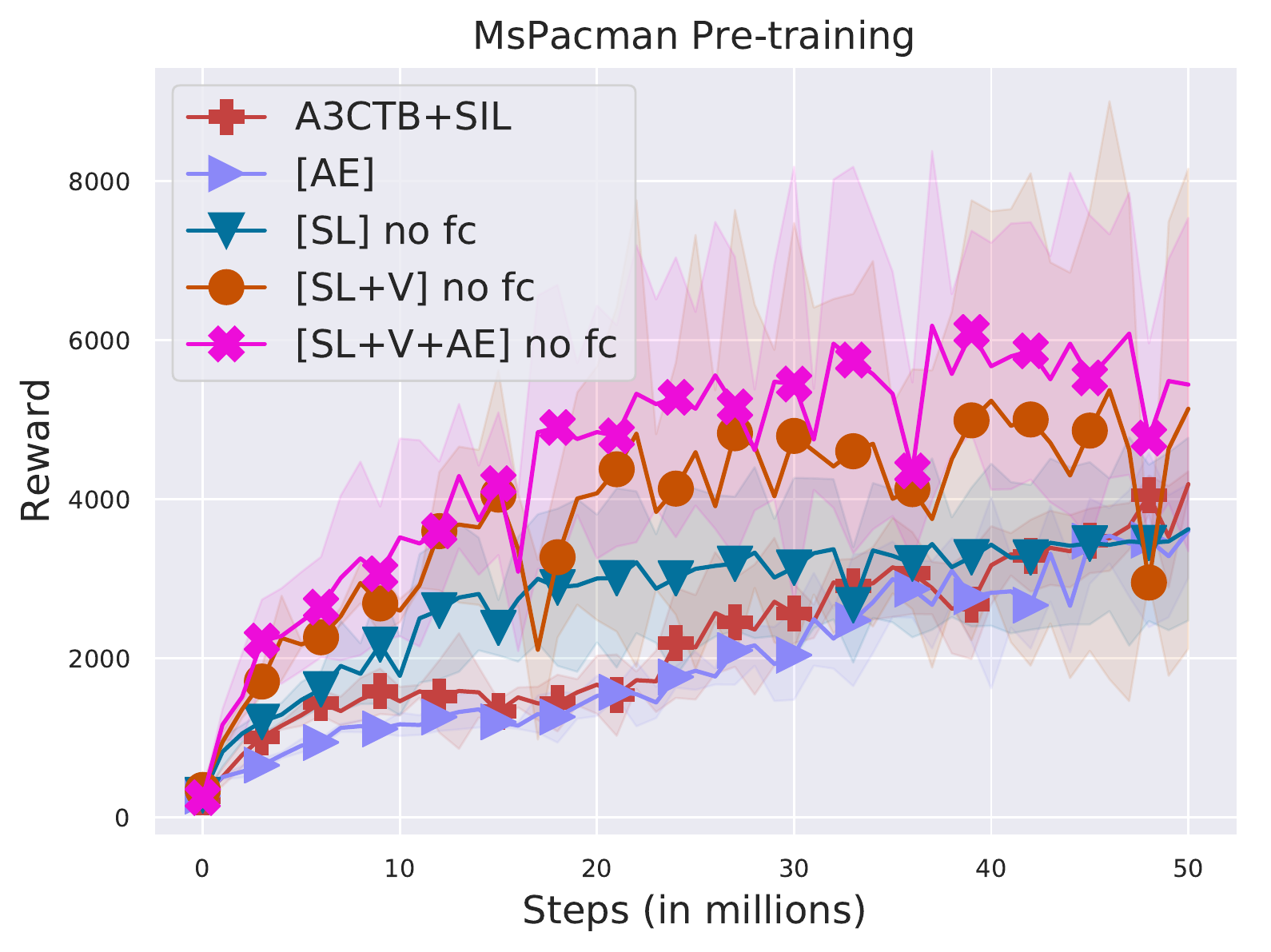}
        \label{fig:pacman-transnofc}
    \end{subfigure}
    \begin{subfigure}[b]{0.49\textwidth}
        \includegraphics[width=\textwidth]{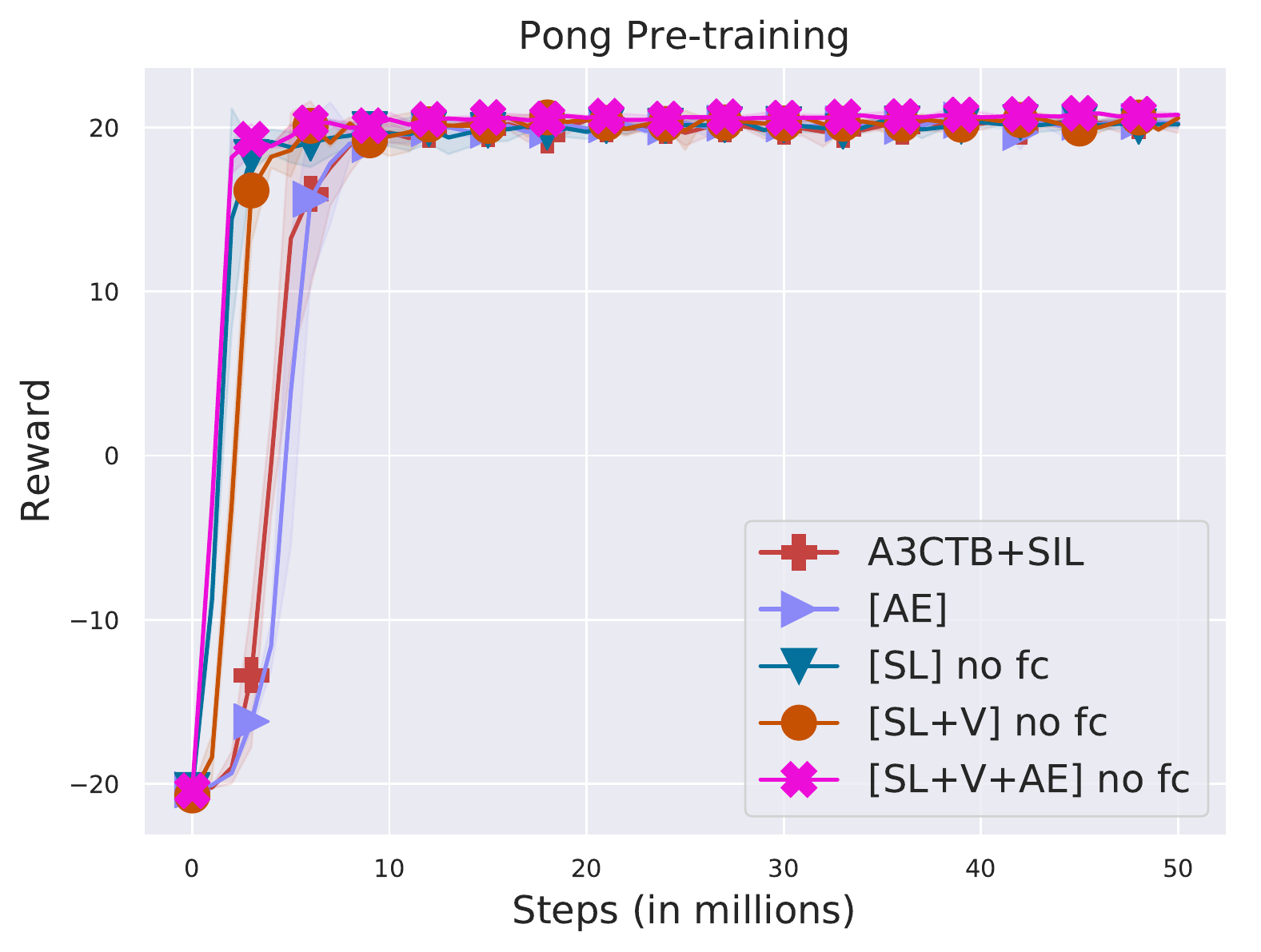}
        \label{fig:pong-transnofc}
    \end{subfigure}
    \caption{Pre-training performance of MsPacman and Pong. Transfer without fully connected output layers (fc2 and fc3). Pre-training methods are shown in brackets. After pre-training, all agents are trained in \emph{A3CTB+SIL}. The x-axis is the total number of training steps (16 actors for methods without SIL; 16 actors plus 1 SIL-learner for methods using SIL). Each step consists of four game frames (frame skip of four). The y-axis is the average testing score over three trials; shaded regions are the standard deviation.}
    \label{fig:transnofc}
\end{figure*}

\subsection{A3CTB+SIL}
 Figure \ref{fig:base} shows the performance of \emph{A3CTB+SIL} when compared to the baseline A3C \cite{a3c}, \emph{A3CTB} \cite{de2018pre}, and \emph{A3C+SIL}. 
\emph{A3C+SIL} helps in both games and shows better improvement in MsPacman. This is consistent with the findings in \citet{sil} that imitating past good experiences encourages exploration, which is beneficial for hard exploration games. Our proposed method \emph{A3CTB+SIL} shows the best performance among all. The largely improved score in MsPacman indicates that it is important for the agent to be able to distinguish big and small rewards (the function of TB); SIL helps to imitate past experiences with large returns.

\subsection{A3CTB+SIL with Pre-training}
\label{sec:result-full}

Since \emph{A3CTB+SIL} has shown the largest improvement without pre-training, from now on, we investigate if using our new pre-training approaches can further accelerate \emph{A3CTB+SIL}. That is, after pre-training, all agents are then trained in \emph{A3CTB+SIL}. Figure \ref{fig:transall} shows the results for pre-training methods. In the game of MsPacman, while \emph{[SL]} already sees slight improvements over the baseline, both \emph{[SL+V]} and \emph{[SL+V+AE]} show superior improvements over \emph{[SL]}, achieving a testing reward (averaged over three trials) at around 8,000. Compared with some state-of-the-art results such as \citet{dqfd} and \citet{sil} where the final rewards for MsPacman are roughly around 5,000 and 4,000 respectively,\footnote{Numbers approximately read from the figures of the mentioned papers.}, our method largely exceeded theirs. 

In the game of Pong, all pre-training methods exceed the baseline performance but the amount of improvements are not as large as in MsPacman. One reason could be that Pong is a relatively easy game to play in Atari and the agent is able to find a good policy even when learning from scratch. In addition, note that \emph{[SL]} actually has the fastest learning speed among other pre-training methods, which is intuitively reasonable. Catching the ball is probably the most important behavior to learn in Pong and this movement is highly associated with the classification of actions; learning the value function and the feature representations did not seem to add additional benefits than learning just an action classifier.

\subsubsection{Ablation Study}
We want to see how useful the feature representations learned during the pre-training stage are to a DRL agent. It is known that general features are retained in the hidden layers of a neural network while the output layer is more task-specific \cite{yosinski2014transferable}. Therefore, in this set of experiments we exclude all output fully connected layers (fc2 and fc3) and only initialize a new A3C network with pre-trained convolutional layers and the fc1 layer, then train it in \emph{A3CTB+SIL}. This experiment will allow us to investigate how important is the general feature learning as to the task-specific policy learning for a DRL agent.

Figure \ref{fig:transnofc} shows the results on transferring pre-trained parameters without the fully connected out layers (``no fc'' refers to ``no fc2 and no fc3''). Note that the \emph{[AE]} pre-training refers to pre-training with only the reconstruction loss $L_{ae}^{saev}$ (see Equation \eqref{eq:saev_loss}); since the autoencoder model trains on fc1 and does not affect fc2 and fc3, we consider it as ``transfer without fc2 and fc3'' and present its results here instead of in previous experiments where ``all layers are transferred.'' It is interesting to observe that, in MsPacman, the performance of \emph{[SL+V] no fc} and \emph{[SL+V+AE] no fc} dropped relatively compared to when transferring all layers. While in Pong, not transferring the output layers did not affect the performance as much. This indicates again that, due to the nature of the game MsPacman, more exploration is needed and only having a good initial feature representation is not as good as when knowing some priors about both the features and the behaviors. However, in games that require less strategy learning and exploration, having a good initial feature of the environment can already provide a performance boost. For example in Pong, even when not transferring the output layers, the information retained in the hidden layers are still highly related to the paddle movement (since it classifies actions), which could be the most important thing to learn in Pong.

The lower performance of not transferring the output layers for MsPacman also shows the benefits of pre-training both the policy layer and the value layer. As shown in Figure~\ref{fig:transall} for MsPacman, when using all layers of the pre-trained network, it has a higher initial testing reward compared to the baseline. This indicates that the initial policy was better after pre-training both policy and value layers. 
We believe that this could be a way of identifying when to use the full pre-trained network and when to exclude the output layers. Future work should study 
if the following hypotheses hold true:
\begin{enumerate}
    \item If the initial performance of the pre-trained network is better than the baseline, then one should use the full pre-trained network.
    \item Otherwise, it might be better to use the pre-trained network without the output layers.
\end{enumerate}

\section{Related Work}\label{sec:relatedwork}

Our work is largely inspired by the literature of transfer for supervised learning \cite{pan2010survey}, particularly in deep supervised learning where a model is rarely trained from scratch. Instead, parameters are usually initialized from a larger pre-trained model (e.g., ImageNet \cite{2015imagenet}) and then trained in the target dataset. \citet{yosinski2014transferable} performed a thorough study on how transferable the learned features at each layer are in a deep convolutional neural network, showing that a pre-trained ImageNet classification model is able to capture general features that are transferable to a new classification task. Similarly, unsupervised pre-training a neural network can extract key variations of the input data, which in term helps the supervised fine-tuning stage that follows to converge faster \cite{erhan2010does,bengio2013representation}. In this work, we combine the supervised and unsupervised methods into the supervised autoencoder framework \cite{sae} as our pre-training stage. Intuitively, the supervised component could guide the autoencoder to learn features that are more specific to the task.

It has been shown that incorporating useful prior information can benefit the policy learning of an RL agent \cite{schaal1997learning}. Learning from Demonstrations (LfD) integrates such a prior by leveraging available expert demonstrations; the demonstration can either be generated by another agent or be collected from human demonstrators \cite{argall2009survey}. Some previous work seeks to learn a good initial policy from the demonstration then later on combine with the agent-generated data to train in RL \cite{kim2013learning,Piot2014BoostedBR,Chemali2015DirectPI}. More recent approaches have considered LfD in the context of DRL. \citet{christiano2017deep} proposes to learn the reward function from human feedback. \citet{gao2018reinforcement} considers the scenario when demonstrations are imperfect and proposes a normalized actor-critic algorithm. Perhaps the closest work to ours is the deep Q-learning from demonstration (DQfD) algorithm \cite{dqfd}. In DQfD, the agent is first pre-trained over the human demonstration data using a combined supervised and temporal difference losses. However, during the DQN training stage, the agent continues to jointly minimize the temporal difference loss with a large margin supervised loss when sampling from expert demonstration data. Our work instead uses a supervised autoencoder as pre-training, which explicitly emphasizes the representation learning and our pre-training losses are not carried through in the RL training. DQfD was further improved as Ape-X DQfD by \citet{apexdqn} where the transformed Bellman operator was applied to reduce the variance of the action-value function and is applied to a large-scale distributed DQN. We empirically observe that our pre-training approaches obtained a higher score in MsPacman than that of DQfD and are comparable to Ape-X DQfD (see Section \ref{sec:result-full}). However, we are unable to compare directly as we do not have the computational resources to run such a large scale experiment. In addition, note that Ape-X DQfD does not have a pre-training stage; DQfD addresses both feature and policy learning while our work only addresses feature learning. Therefore, our method is not directly comparable to the above. 

The use of unsupervised auxiliary losses has been explored in both deep supervised learning and DRL. For example, \citet{zhang2016augmenting} uses unsupervised reconstruction losses to aid in learning large-scale classification tasks; \citet{unreal} combines additional control tasks that predict feature changes as auxiliaries. Our methods of pre-training via supervised autoencoder can be viewed as leveraging the reconstruction loss as an auxiliary task, which guides the agent to learn desirable features for the given task. 

\section{Discussion and Conclusion}\label{sec:conclusion}
In this work, we studied several pre-training methods and showed that the A3C algorithm could be sped up by pre-training on a small set of non-expert human demonstration data. In particular, we proposed to integrate rewards in supervised policy pre-training, which helps to quantify how good a demonstrated action was. The component of the value function and autoencoder pre-training yielded the most significant performance improvements and exceeded the state-of-the-art results in the game of MsPacman. Our approach is light-weight and easy to implement in a single machine. 

While pre-training works well in the two games presented in this paper, there is a need to perform this experiment in more games to show the generality of our method. Looking into what features are learned during pre-training is also interesting to study. \citet{de2018pre} visualized the feature learned in supervised pre-training and a final DRL model and found that they share some common patterns, indicating why pre-training is useful. In future work, we are interested in studying how the learned feature pattern differs from each pre-training method. Lastly, pre-training methods only address the problem of feature learning in DRL but do not aid policy learning. To further accelerate learning, we plan on looking into how could policy learning be improved using human demonstration data. Some existing work like DQfD integrate the supervised loss not only during pre-training but also during training the DRL agent \cite{dqfd}; others leverage human demonstrations as advice and constantly providing suggestions during policy learning \cite{wang2017improving}. We attribute these as our future directions.  

\begin{acks}
The A3C implementation was a modification of \url{https://github.com/miyosuda/async_deep_reinforce}. The authors thank NVidia for donating a graphics card used in these experiments. This research used resources of Kamiak, Washington State University's high performance computing cluster, where we ran all our experiments.
\end{acks}


\bibliographystyle{ACM-Reference-Format}  
\bibliography{arxiv-bibliography}  

\end{document}